# A Study and Analysis of a Feature Subset Selection Technique using Penguin Search Optimization Algorithm (FS-PeSOA)

Agnip Dasgupta[1], Ardhendu Banerjee[2], Aniket Ghosh Dastidar[3], Antara Barman[4], Sanjay Chakraborty[5]

[1,2,3,4,5] *Information Technology Department, TechnoIndia, Saltlake, Kolkata, India*

*Abstract*

*In today's world of enormous amounts of data, it is very important to effectively extract knowledge from it. This can be accomplished by feature subset selection. Feature subset selection is the method of selecting a minimum number of features with the help of which our machine can learn and accurately predict which class a particular data belongs to. In this book chapter, we will introduce a new adaptive algorithm called Feature selection Penguin Search optimization algorithm which is a meta-heuristic feature subset selection method. It is adapted from the natural hunting strategy of penguins, in which a group of penguins take jumps at random depths and come back and share the status of food availability with other penguins, and in this way, the global optimum solution is found namely Penguin Search Optimization Algorithm. It will be combined with different classifiers to find the optimal feature subset. In order to explore the feature subset candidates, the bio-inspired approach Penguin Search optimization algorithm generates during the process a trial feature subset and estimates its fitness value by using three different classifiers for each case: Random Forest, Nearest Neighbour and Support Vector Machines. However, we are planning to implement our proposed approach Feature selection Penguin Search optimization algorithm on some well-known benchmark datasets collected from the UCI repository and also try to evaluate and compare its classification accuracy with some state of art algorithms.*

## 1. Introduction

Machine Learning is a branch of Artificial Intelligence (AI), which allows applications to become more authentic in anticipating to which class a particular data resides. There are various applications of machine learning which spread over the areas like healthcare, finance, retail, travel, Social Media, advertisements and most importantly data mining[1]. There are two types of learning: Supervised Learning and Unsupervised Learning. In machine learning, a feature is an alone measurable property or trait of a phenomenon being observed. Feature subset selection is one of the important tasks in data mining. To perform effective data mining and pattern recognition tasks, we need to take the help of an efficient feature selection technique. However, due to a large number of features present in high-dimensional datasets, there is a chance of unnecessary overfitting which increases the overall computational complexity and reduces the prediction accuracy of the procedure. This feature selection problem belongs to the set of NP-hard problems where the complexity increases exponentially if the number of features along with the size of datasets is increasing. Feature subset selection not only helps us get rid of the curse of dimensionality but also helps us shortening the training time and simplifying the model making it easier for the analysts to interpret it. There are various approaches which deal with both supervised and unsupervised ways of feature subset selection[2][3][4]. There are two classes of feature selection methods, such as 1) filter based feature selection and 2) wrapper-based feature selection.

Filter methods are generally a part of the pre-processing step. Here, each feature is selected on the basis of their scores in various statistical tests, some of which are Pearson's Correlation (PCA), Linear Discriminant Analysis(LDA), Analysis of variance(ANOVA), and Chi-Square. The other methods are wrapper methods where a set of a subset is randomly chosen and then the efficiency is checked, after that other features apart from the subset are chosen and the results are checked again, or some irrelevant or less important features are removed from the subset. This continues until we find the ideal subset. In wrapper methods, the problem is reduced to a search problem. Using wrapper methods are quite expensive [5]. Some of the examples of wrapper methods are Forward selection, backward elimination, and Recursive Feature Elimination. Apart from these two methods, we can also use an embedded method which includes the qualities of both Wrapper methods and Filter methods. It is used by those algorithms which have built-in feature subset selection methods[6].

After feature subset selection, we use classifiers to classify which class the particular data belongs to. However, we have tested our algorithm using KNN (K-Nearest Neighbours), SVM (Support Vector Machine) and Random Forest classifiers[7]. KNN is also called a K-Nearest Neighbours, is a very simple non-parametric decision procedure that is used to assign unclassified observations a class with the use of a training data set[8].

In this chapter, we have worked upon Feature subset selection using 'Penguin Search Algorithm', inspired by the hunting strategy of the Penguins[9]. Also, we have tested our algorithm on some popular UCI data sets including Iris, Pima, Wisconsin etc. Then we have compared our work with the existing feature subset algorithms. We have also used three different types of classifiers and checked how good our algorithm works in terms of parameters such as Accuracy, Precision, Recall, and F1 score. There are several existing algorithms or approaches for 'Feature subset selection', which are nature inspired like Ant, Bee Colony, Whale, etc., and so we are using 'Penguin Search Optimization Technique (PeSOA)' in our work, which is based on the way penguins hunt for food[9]. Penguins take random jumps in different places and random depths to find out fishes after they find fishes they come back and communicate the food availability with the other penguins, this continues till the penguins find the best place or the place where maximum fishes are present or the global maxima. The goal of finding global maxima continues until the oxygen level of penguins does not get depleted. Each penguin has to return back to the surface after each trip. The duration of the trip is measured by the amount of oxygen reserves of the penguins and the speed at which they use it up, or their metabolism rate. This behaviour of the penguins has given us motivation for the development of a new optimization method based on this strategy of penguins. Penguins are sea birds, and they are unable to fly[9]. Meta-heuristics, which is mainly used for the development of new artificial systems and it is effective in solving NP-hard problems. It can be classified in various ways. The first work in the field of optimization commenced in the year 1952 based on the use of stochastic manner. Rechenberg diagrammed the first algorithm using evolution strategies in the optimization in 1965[9]. Most of the methods utilize the thought of population, in which a set of solutions are calculated in parallel at each iteration, such as genetic algorithms, PSO (Particle Swarm Optimization Algorithm) and the ACO (Ant Colony Optimization Algorithm)[10][11]. Other meta-heuristic algorithms use the search results based on their past experiences in order to guide the optimization in following iterations by putting a learning stage of intermediate results that will conduct to an optimal solution. And our chapter related to "Penguins Search Optimization Algorithm (PeSOA)" is an example of it. It mainly depends on the collaborative hunting strategy of penguins. The catholic optimization process starts based on the individual hunting process of each penguin, who must shares information to his group related to the number of fish found of their individual hunting area. The main objective of the group conversation is to achieve a global solution (the area having abundant amounts of food). The universal solution is chosen by the selection of the best group of penguins who ate the maximum number of fish. Comparative studies based on other meta-heuristics have proved that PeSOA accomplishes better answers related to new optimization strategy of collaborative and dynamic research of the space solutions.

This rest of the book chapter is organized as follows. A brief literature review has been done in section 2. In section 3, we have described our proposed work with a suitable flowchart diagram. Then we have described a detailed performance analysis of our proposed approach in section 4. We have also compared our proposed work with some previous studies related to different parameters of classification in section 4 and finally, section 5 describes the conclusion of this chapter.

## 2. Literature Review

In the last decade, multiple researchers adopted various optimization methods for solving the feature selection problem. In one of the earliest works, a novel marker gene feature selection approach was introduced. In this approach, a few high graded informative genes were elected by the signal-noise ratio estimation process. Then a novel discrete PSO algorithm was used to choose a few marker genes and SVM was used as an evaluator for getting excellent prediction performance on Colon tumour dataset. The authors have introduced an algorithm called a swarm intelligence feature selection algorithm which is mainly based on the initialization and update of the swarm particles. In their learning process, they had tested the algorithm in 11 microarray datasets for brain, leukaemia, lung, prostate etc. And they have noticed that the proposed algorithm was successfully increasing the classification accuracy and reduce the number of chosen features compared to other swarm intelligence process. The authors have compared the utilization of a PSO and a Genetic Algorithm (GA) (both illuminated with SVM) for the classification of high dimensional microarray data. Those algorithms are mainly used for finding small samples of informative genes amongst thousands of them. An SVM classifier with 10- fold cross-validation was applied in order to validate and assess the provided solutions [12]. There is one paper where Whale Optimization Algorithm (WOA) was introduced through which a new wrapper feature selection approach is proposed. And it is a recently proposed algorithm which has not been systematically used to feature selection problems. There are mainly two binary variants of the WOA algorithm were introduced to find the subsets of the optimal feature for classification purposes. In the first case, the aim was mainly to study the impact of using the Tournament and Roulette Wheel selection mechanisms instead of using a random operator in the searching procedure. In a second way, crossover and mutation operators are used to prolong the exploitation of the WOA algorithm. The proposed methods are tested based on standard benchmark datasets. That paper also considers a comprehensive study related to the parameter setting for the proposed technique. The results display the ability of the proposed approaches in searching for the optimal feature subsets [13]. In a paper, a noble version of the binary gravitational search algorithm (BGSA) was proposed and used as a tool to choose the best subset of features with the objective of improving classification accuracy. The BGSA has the ability to overcome the stagnation situation by enhancing the transfer function. And then the search algorithm was used to explore a larger group of possibilities and avoid stagnation. To assess the proposed improved BGSA (IBGSA), classification of some familiar datasets and improving the accuracy of CBIR systems were experienced. And the results were compared with those of original BGSA, genetic algorithm (GA), binary particle swarm optimization (BPSO), and electromagnetic-like mechanism. Relative results ensured the effectiveness of the proposed IBGSA in feature selection [14]. Various optimization techniques(Hill-climbing, Particle swarm optimization etc.) are used to do an efficient and effective unsupervised feature subset selection [6].

It can be seen from a survey of existing work that a few researchers in the last decade have tried to solve the problem of feature selection using optimization techniques. There have been couples of attempts by researchers to unify diverse algorithms for supervised and unsupervised feature selection. However, we think PeSOA can provide better selection strategy in case of wrapper based feature selection process. In our recent work, we discuss this area of feature selection.

### 3. Proposed Work

In our algorithm, each location where the penguin jumps into is considered a feature or a dimension, and as the penguins dive to find the food, the penguins dive deep into the feature to check whether that particular feature is important or not or the amount of food or fishes is ample or not. The results are shared among the other penguins once they come to the surface and this search for finding the optimum result or the minimum number of features by which classification gives good results continues till the oxygen level depletes or the best features or best positions are found out by the penguins i.e. [15], in terms of penguins places with the highest amount of fishes or food. On the basis of these things, we have designed our algorithm.

**3.1 Pseudo Code of the Proposed FS-PeSOA Algorithm**

*FS-PeSOA (Feature Selection using Penguin Search Optimization Algorithm)*

*Start;*

*Obtain Dataset;*

*Split the dataset in the ratio of 80:20 as training set and test set;*

*Generate Random population of P penguins;*

*Initialize Oxygen reserve for Penguins*

*Initialize the first location of Penguins*

*While (iterations<Oxygen) do*

    *For each penguin j do*

        *Look for Fish available (Calculate fitness of available data for the current Penguin with the help of Fitness Function 4.3.2).*

        *Determine the quantity of fish available.*

        *Update the Position of the Penguin (based on the Position Update Logic 4.3.3)*

    *End for;*

    *Update the best solution;*

    *Get the food quantity (fitness) data from the penguins to update the group.*

    *Scale the food quantity (fitness) data for position update in next iteration.*

    *Update the Oxygen reserve for the penguins (using Oxygen Update Logic 4.3.4).*

*End While;*

*Find out the features that qualify the fitness cutoff*

*Save the obtained feature subset;*

*Use this subset of features to undergo classification using SVM (Support Vector Machines), KNN(K- nearest neighbor) and Random Forest;*

*Performance analysis using precision, recall, f-score, and accuracy;*

*End;*

The first step of the Machine Learning is choosing the dataset we intend to work upon. These datasets have been explained in Table 1. After the dataset has been chosen the data needs to be normalized or scaled to a particular range. This is done because it might be possible that one attribute ranges between 1-100 and the other attribute ranges from 10,000 to50,000. This type of variance in the dataset will be affecting our results. Hence, we need to scale it down to a particular range setting the upper limit and the lower limit. After the scaling has been done, the dataset is divided into two parts- training dataset and testing dataset. This is generally in the ratios of 70:30 or 80:20. Training dataset, as the name suggests is used to train our Machine and the testing dataset is used to test our algorithm, how efficiently it works. This step is also called as data-splitting. The overall workflow of our proposed approach is shown in Fig.1.

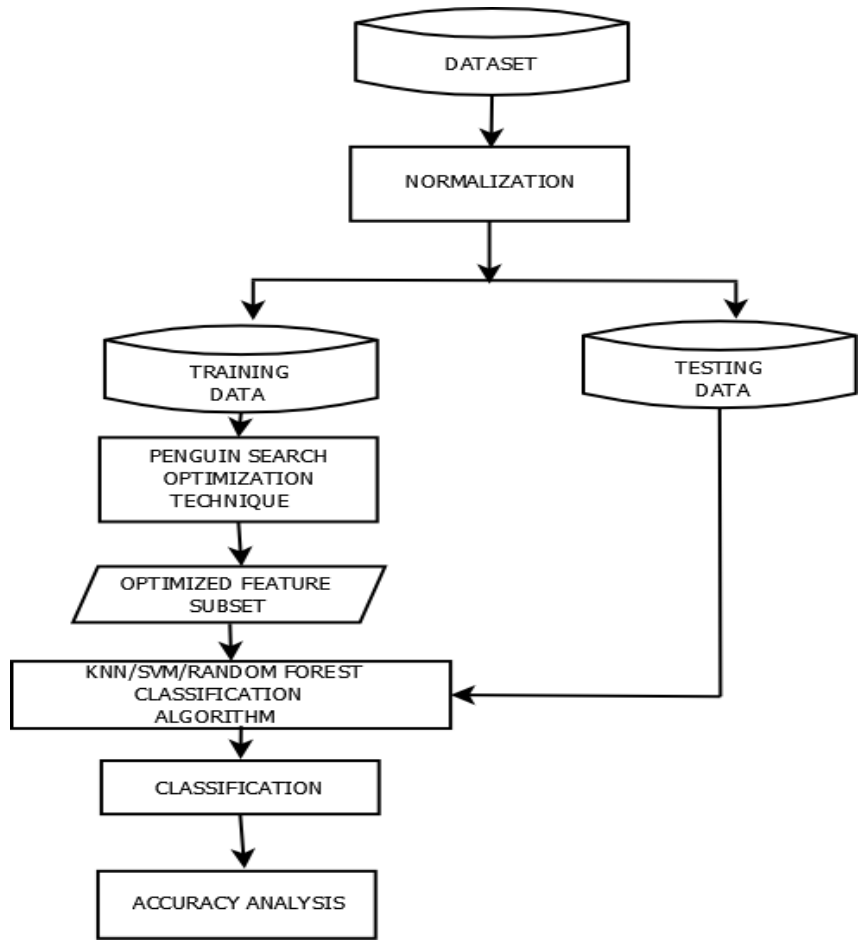

Fig.1 The overall flowchart of the proposed FS-PeSOA algorithm

Now, after the data gets split into two parts, we will execute our proposed algorithm for feature subset selection and the best features are selected out. This step is known as Feature subset selection which has been explained in the Introduction part. Briefly, it is selecting the minimum number of features by which our Machine can identify which class a particular data belongs to. After Feature subset selection has been done, several classifiers are used like SVM, KNN, Random Forest etc. for the further classification process. However, after the completion of the training phase, we would like to test our algorithm with the testing dataset which we kept aside and will be performing the same steps again of classification.

After the classification is done, we are checking how accurate our results are and how well it performs as compared to other benchmark feature selection algorithms.

In the 'Result Analysis' section, the comparison factors are accuracy, precision, recall and F1 score and the benchmark feature selection algorithms which we have compared are LDA and PCA.

### 3.3. Discussion
### 3.3.1. Hunting Strategy of Penguins

The hunting strategy of the penguins is theorized by [9]. This hypothesis theorized that their hunting strategy may be explained in an economical manner. The penguins are a biological being which have a definite amount of capacity, by capacity it means that they have a definite amount of oxygen and strength to fulfill their search for food. While the process of finding the food for their survival they tend to dive in water for finding fishes. They go economical where they need to find food, the point is they go for a search in food where the amount of food found is on par with the energy spend to find the food. Moreover, they have a limited amount of oxygen left with them that limits the amount of time they can continue on with the total hunting procedure. They do dive in the water and look for the available food and consume them, and when they come back to the surface again they communicate with the whole group about the amount of food found and at which location, this communication among the penguins plays a vital role in this whole searching of food by the penguins. Whenever they communicate among them about the data of the location and quantity of food found the penguins that have found a lesser amount of food tend to travel to the location which has been reported to have more amount of food in comparison to the others. However, we have tried to visualize this hunting strategy of the penguins to optimize our searching technique for an optimal subset of all the features of the whole dataset in this chapter. We have tried to visualize the amount of food found by the penguins as the goodness or the fitness of the features the penguins travel to. All the penguins work as a whole unit to deliver a single objective to find the maximum amount of food that they can find with a limited number of resources. There is a cycle where they go in search of food and come back with food and communicate with the other penguins with the data of the food and location and then the penguins travel to other locations and then they again go in search of food. This whole cycle goes on continuing until and unless they have found the required amount of food.

### 3.3.2. Fitness Function Evaluation

In contrary to the actual world phenomenon of finding food by the penguins they look for real-time quantity of food and based on this quantity of food they compare the quality of a particular location. For our datasets to find the quality or the fitness of the features with their set of observations we have used the Eigen Vector Function as described in [16].The Eigenvalue works on the principle of calculating the variance of the particular feature with the total number of records for the particular feature. The basic idea is that the more the values of a particular feature scattered in a more varied range, the feature is fit for being used for the purpose of classification. Whenever a penguin goes to a particular feature the Eigenvalue function is used to get the variance of the feature and this is the fitness. Based on this fitness value the features are selected on the cut-off criteria. Now, suppose we have a data set of {x(i),i=1,2,3,…,m} of m different features . So the data lies in an m dimensional subspace and the data basically lies between the diagonal of this m dimensional subspace.  Now we find the mean of the data by the equation (1) given below[16].

$$\mu = \frac{1}{m}\sum_{i=1}^{m} x^{(i)}$$

........................................ (1)

Replace the x $^{(i)}$ data with x $^{(i)}$_μ and from this, we get the mean of the data normalized. However, the data with zero and no mean is omitted and normalized. Now we find the variance of the data with the equation (2).

$$\sigma_j^2 = \frac{1}{m}\sum_i (x_j^{(i)})^2$$

........................................ (2)

Replace x $^{(i)}$ with X $^{(i)}$/σ, from this we get the data scaled in a particular range and this normalizes the data with respect to the whole dataset and all features. This is the final step of this normalization. Now that we get the variance of the data that we have got[16].

This resultant variance is the score of the feature with the set of all the data records. This represents how the data is scattered in the maximum to minimum scale for the feature, which will determine the features which should be suitable for classification.

### 3.3.3. Position Update Logic

As per the penguins will go towards the location that would have a better food quantity than the others. Here the penguins would initially go to a feature respective to that fitness value acquired f(x$^{(i)}$) which is in a scale of 0 to 1. This fitness value multiplied with m, where m is the total number of features we get the new position for our penguin for the next iteration.

$$\text{Position, } (n+1) = f(x\ (i))*m \quad\quad \ldots\ldots\ldots\ldots\ldots\ldots\ldots\ldots\ldots\ldots(3)$$

Based on this position update logic the penguins will keep updating their position until they find the optimal amount of food.

### 3.3.4. Oxygen Update Logic

For the penguins, the quantity of oxygen is a limiting factor that restricts them to continue the whole searching procedure for an infinite amount of time. The oxygen is a physical quantity that is basically limited to all beings. Whenever their oxygen reserve is depleted they return back and end the search for food. In our approach of Penguin Search Optimization, we have used the idea of oxygen to limit the number of iterations the penguins will go until they stop to find out the food that they have. We have a predefined value of the oxygen that is same for all the penguins in the beginning and after each iteration, the amount of oxygen gets reduced by a fixed value that after a definite amount of time extinguishes to bring an end to the series of iterations. This can also be considered as the number of generation our iteration will go on until an optimal solution is reached.

## 4. Result Analysis

The implementation of the algorithm has been done on Python programming language using Anaconda as the software and on Jupiter notebook and the experiments are tested on a computer with system specifications of 4 GB RAM, Intel i3 core Processor, and 500GB Hard Disk memory. Seven real-world 'UCI Machine Learning Repository' approved datasets are used to assess the efficiency of our proposed algorithm[17]. Some of them have about 4-5 features and some have about 30-40 features which make it appropriate for us to perform the Feature subset selection. These datasets are also diverse in context of the number of classes and samples. These datasets have been represented in Table.1.

These datasets include Ion, Pima, Iris, Vehicle, Wisconsin, Glass and Wine. Using our proposed algorithm i.e., FS-PeSOA, we have performed the Feature subset selection and then using KNN, Random Forest and SVM classifiers to perform the classification. We have also tested the datasets using other algorithms like PCA and LDA and we will compare the results hoping to get better performance.

Table 1.List of datasets from UCI [17] which are tested using PESOA

| Datasets | No of Observations | No of Classes |
| --- | --- | --- |
| Iris | 150 | 3 |
| Glass | 214 | 6 |
| Ion | 351 | 2 |
| Pima | 768 | 2 |
| Vehicle | 846 | 4 |
| Wine | 178 | 3 |
| Wisconsin | 569 | 2 |

These are the UCI approved datasets which we have been used to check the efficiency of our algorithm. According to the UCI repository of machine learning[17], Iris is the flower dataset which is perhaps the best-known database to be found in the pattern recognition field. It has 3 classes and 4 attributes, and the attributes include sepal length, petal length, sepal width, and petal width. The classes are Iris Setosa, Iris Versicolour, and Iris Virginica.

Ion dataset or ionosphere dataset is a collection of radar data collected in Goose Bay, Labrador. The system has a phased array of 16 high-frequency antennas with a transmitting power on the order of 6.4 Kilowatts and the targets are free electrons which are present in the Ionosphere. This dataset consists of 33 attributes and two classes namely good radar and bad radar.

Good radar is that which shows some structure in the Ionosphere and the Bad is that whose signals cannot pass through the Ionosphere. Pima dataset is collected by a survey done on 768 people by National Institute of Diabetes and Digestive and Kidney Diseases. The motive of the dataset or the machine learning part of the dataset is to find out whether a particular patient has diabetes or not on the basis of various diagnostic measurements. In the dataset, the survey is done on Indian females who are above the age of 21 years and are of Pima Indian Heritage. This dataset has two classes, either the patient is suffering or not. According to the 'UCI repository of Machine Learning', the wine dataset contains the results of a chemical analysis of wines. It has 3 classes and multiple attributes like fixed acidity, volatile acidity, citric acid, residual sugar, chlorides, free sculpture dioxide, total sculpture dioxide, density, pH, sulphates, alcohol etc. The number of features before subset selection was 13 and after feature subset selection it got reduced to 4.

According to UCI Repository, the Vehicle dataset classifies a given silhouette as one of four types of vehicles with 18 features some of which are Compactness, Circularity, Radius Ratio, Elongatedness etc. This data was originally gathered in the year1986-1987 by JP Siebert. According to the UCI Repository of Machine Learning, the study of classification of types of glass is motivated by a criminological investigation. At the scene of the crime, the glass left can be used as evidence. It has 6 classes and multiple attributes like Id number: 1 to 214, refractive index, Sodium (unit measurement: weight percent in corresponding oxide, as are attributes 4-10), Magnesium, Aluminium, Silicon, Potassium, Calcium, Barium, Iron, Type of glass: (class attribute): building_windows_float_processed,building_windows_non_float_processed,vehicle_windows_float_pro cessed, vehicle_windows_non_float_processed (none in this database), containers, tableware, headlamps. The number of features before subset selection was 10 and after feature subset selection it got reduced to 4. According to the UCI Repository of Machine Learning, the features are computed from a digitized image of a fine needle aspirate (FNA) of a breast mass. They describe the characteristics of the cell nuclei present in the image. The Wisconsin data set has 2 classes and multiple attributes like ID number,

Diagnosis (M = malignant, B = benign), and ten real-valued features are computed for each cell nucleus like radius (mean of distances from center to points on the perimeter), texture (standard deviation of grayscale values), perimeter, area, smoothness (local variation in radius lengths), compactness (perimeter^2 / area - 1.0), concavity (severity of concave portions of the contour), concave points (number of concave portions of the contour), symmetry, fractal dimension ("coastline approximation" - 1).The number of features before subset selection was 30 and after feature subset selection it got reduced to 6.Table2 introduces us to the results i.e., Accuracy, Precision, Recall and F1 score of our algorithm when tested upon the datasets in Table 1. The table also explains the total number of features which were present before the Feature subset selection and after that was done. The table gives detailed information or results when different classifiers have been used.

Table 2 Proposed algorithm result analysis

| Dataset | Total Feature | | Classification | Accuracy | Precision | Recall | F1 Score |
|---|---|---|---|---|---|---|---|
| | Before | After | | | | | |
| Iris | 4 | 1 | KNN | 90 | 0.94 | 0.90 | 0.90 |
| | 4 | 1 | Random forest | 90 | 0.94 | 0.90 | 0.90 |
| | 4 | 1 | SVM | 90 | 0.94 | 0.90 | 0.90 |
| Glass | 10 | 4 | KNN | 81.39 | 0.82 | 0.81 | 0.80 |
| | 10 | 4 | Random forest | 79.06 | 0.81 | 0.79 | 0.79 |
| | 10 | 4 | SVM | 81.39 | 0.80 | 0.81 | 0.79 |
| Ion | 33 | 8 | KNN | 95.91 | 0.88 | 0.86 | 0.85 |
| | 33 | 8 | Random forest | 91.54 | 0.92 | 0.92 | 0.91 |
| | 33 | 8 | SVM | 91.54 | 0.93 | 0.92 | 0.91 |
| Pima | 8 | 1 | KNN | 59.74 | 0.58 | 0.60 | 0.59 |
| | 8 | 1 | Random forest | 64.28 | 0.64 | 0.64 | 0.64 |
| | 8 | 1 | SVM | 65.58 | 0.62 | 0.66 | 0.60 |
| Vehicle | 18 | 5 | KNN | 57.05 | 0.58 | 0.57 | 0.57 |
| | 18 | 5 | Random forest | 52.35 | 0.52 | 0.52 | 0.52 |
| | 18 | 5 | SVM | 58.82 | 0.59 | 0.59 | 0.58 |
| Wine | 13 | 4 | KNN | 69.44 | 0.69 | 0.69 | 0.69 |
| | 13 | 4 | Random forest | 55.55 | 0.64 | 0.56 | 0.57 |
| | 13 | 4 | SVM | 38.88 | 0.22 | 0.39 | 0.28 |
| Wisconsin | 30 | 6 | KNN | 79.04 | 0.83 | 0.82 | 0.84 |
| | 30 | 6 | Random forest | 84.61 | 0.79 | 0.81 | 0.82 |
| | 30 | 6 | SVM | 95.61 | 0.96 | 0.96 | 0.96 |

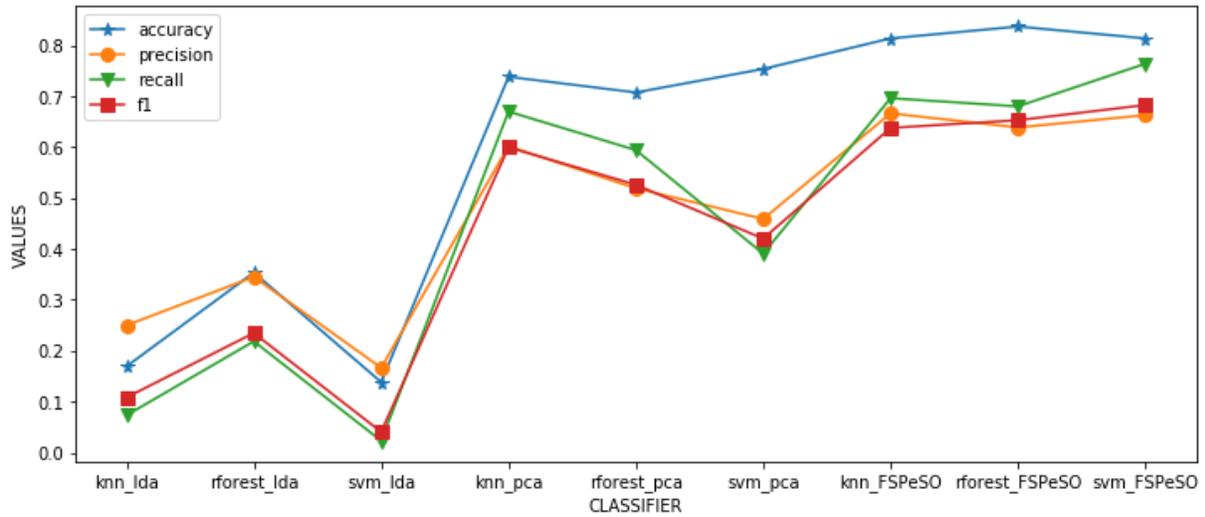

Fig.2 Performance analysis of the proposed FS-PeSOA with other algorithms on Glass dataset

The above Fig.2 shows the performance of accuracy, precision, recall and the F1 score of Glass dataset under KNN, Random Forest and SVM when previously operated by PCA, LDA, and our proposed FS-PeSOA algorithm. The following graph is of Ion dataset and its results.

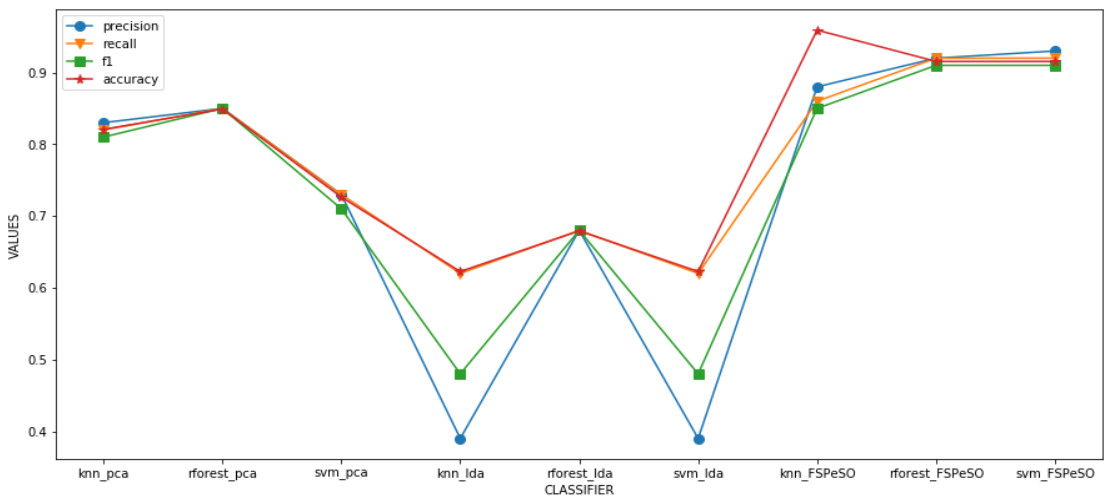

Fig.3 Performance analysis of the proposed FS-PeSOA with other algorithms on Ion dataset

This Fig.3 shows the performance of accuracy, precision, recall and the F1 score of Ion dataset under KNN, Random Forest and SVM when previously operated by PCA, LDA, and our proposed FS-PeSOA algorithm. The following graph is of Pima dataset and its results.

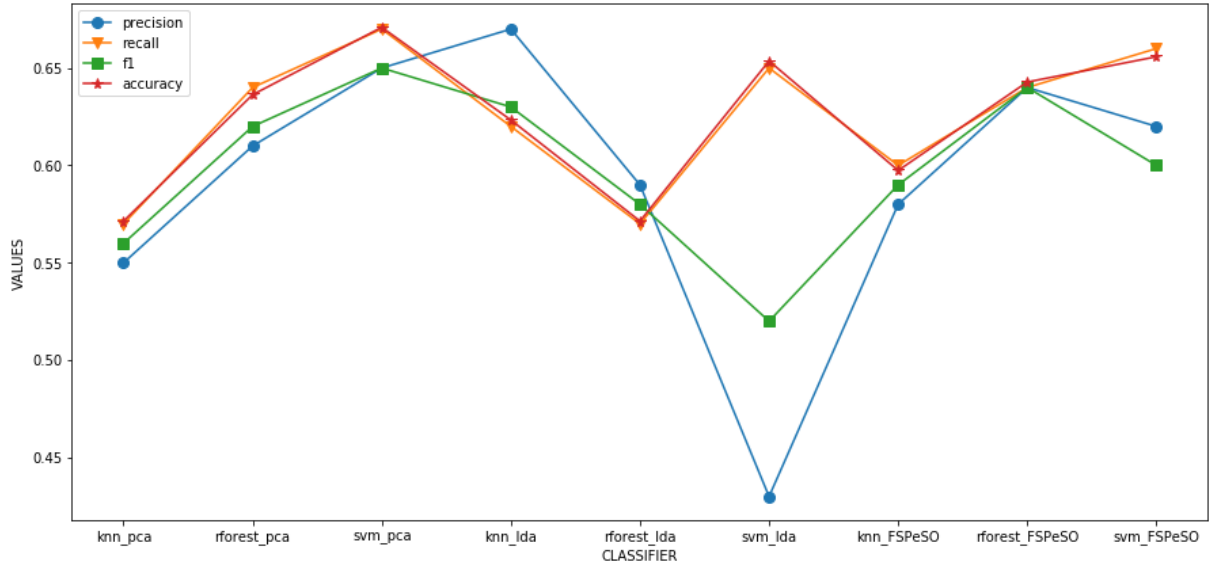

Fig.4 Performance analysis of the proposed FS-PeSOA with other algorithms on Pima dataset

ThisFig.4 shows the performance of accuracy, precision, recall and the F1 score of Pima dataset under KNN, Random Forest and SVM when previously operated by PCA, LDA, and our proposed FS-PeSOA algorithm. The following graph is of Wine dataset and its results.

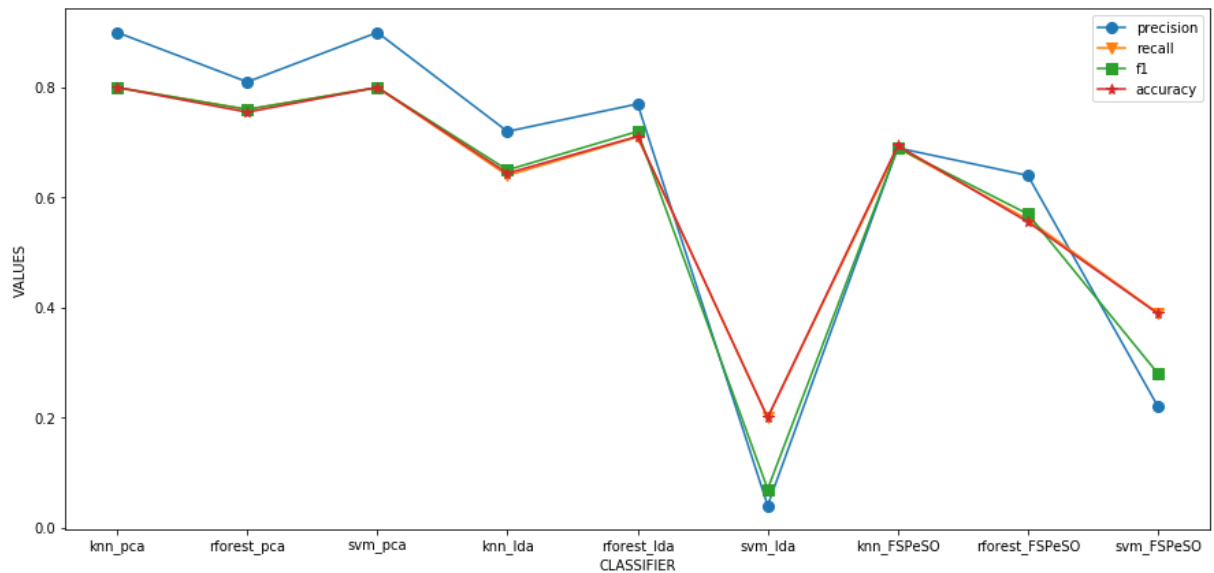

Fig.5 Performance analysis of the proposed FS-PeSOA with other algorithms on Wine dataset

The above Fig.5 shows the performance of accuracy, precision, recall and the F1 score of Wine dataset under KNN, Random Forest and SVM when previously operated by PCA, LDA, and our proposed FS-PeSOA algorithm. The following graph is of Wisconsin dataset and its results.

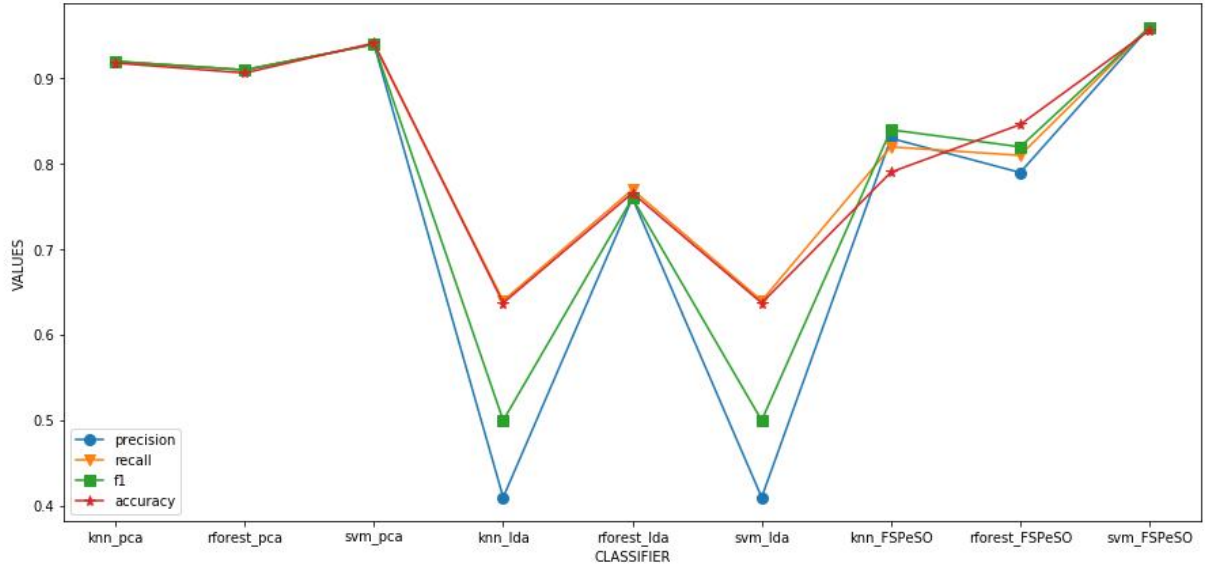

Fig.6 Performance analysis of the proposed FS-PeSOA with other algorithms on Wisconsin dataset

The above Fig.6 shows the performance of accuracy, precision, recall and the F1 score of Wisconsin dataset under KNN, Random Forest and SVM when previously operated by PCA, LDA, and our proposed FS-PeSOA algorithm. The following graph is of Vehicle dataset and its results.

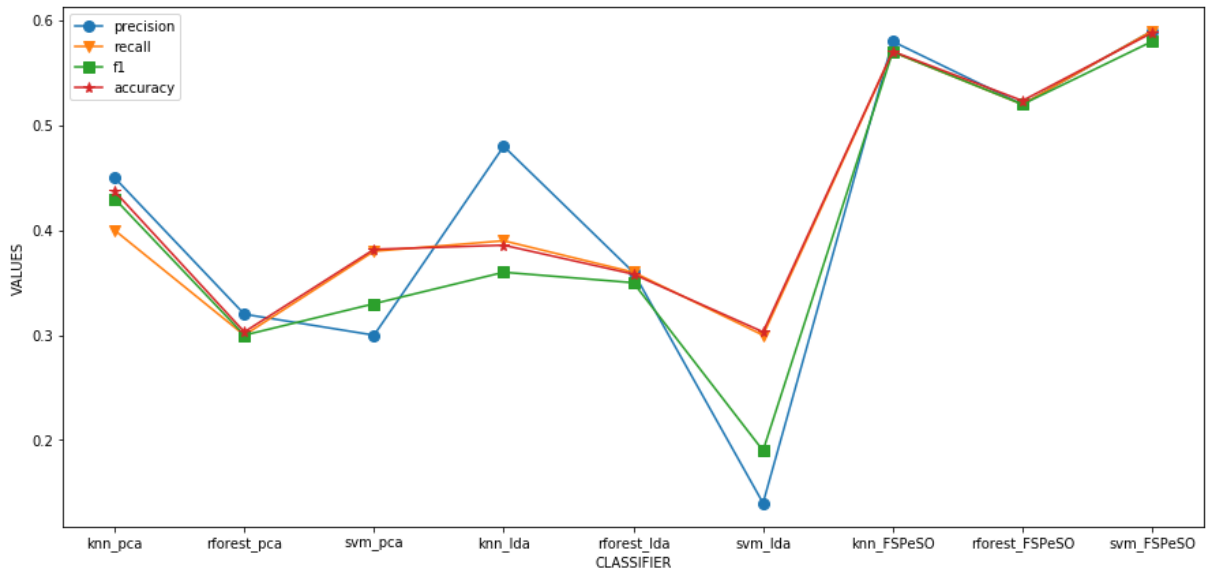

Fig.7 Performance analysis of the proposed FS-PeSOA with other algorithms on Vehicle dataset

The above Fig.7 shows the performance of accuracy, precision, recall and the F1 score of Vehicle dataset under KNN, Random Forest and SVM when previously operated by PCA, LDA, and our proposed FS-

PeSOA algorithm. These are the comparison graphs of different algorithms and our FS-PeSOA algorithm. We have gathered a comparison result in Table 3 which gives a data representation of these graphs.

Table 3a. Comparison based on accuracy, precision, recall and F1-score parameters

| Dataset | Algorithm | Accuracy | Precision | Recall | F1 Score |
|---|---|---|---|---|---|
| Iris | KNN with PCA | 80 | 0.90 | 0.80 | 0.80 |
| | KNN with LDA | 60 | 0.47 | 0.60 | 0.50 |
| | KNN with FSPeSOA | **90** | 0.94 | 0.90 | 0.90 |
| | Random forest with PCA | 77.77 | 0.89 | 0.78 | 0.77 |
| | Random Forest with LDA | 60 | 0.47 | 0.60 | 0.50 |
| | Random Forest with FSPeSOA | **91.22** | 0.94 | 0.90 | 0.90 |
| | SVM with PCA | 80 | 0.90 | 0.80 | 0.80 |
| | SVM with LDA | 60 | 0.47 | 0.60 | 0.50 |
| | SVM with FSPeSOA | **90** | 0.94 | 0.90 | 0.90 |
| Glass | KNN with PCA | 64.61 | 0.62 | 0.65 | 0.62 |
| | KNN with LDA | 53.84 | 0.67 | 0.54 | 0.59 |
| | KNN with FSPeSOA | **81.39** | 0.82 | 0.81 | 0.80 |
| | Random forest with PCA | 66.15 | 0.66 | 0.66 | 0.65 |
| | Random Forest with LDA | 55.38 | 0.60 | 0.55 | 0.56 |
| | Random Forest with FSPeSOA | **79.06** | 0.81 | 0.79 | 0.79 |
| | SVM with PCA | 69.23 | 0.62 | 0.69 | 0.64 |
| | SVM with LDA | 44.61 | 0.22 | 0.45 | 0.29 |
| | SVM with FSPeSOA | **81.39** | 0.80 | 0.81 | 0.79 |
| Ion | KNN with PCA | 82.07 | 0.83 | 0.82 | 0.81 |
| | KNN with LDA | 62.26 | 0.39 | 0.62 | 0.48 |
| | KNN with FSPeSOA | **95.91** | 0.88 | 0.86 | 0.85 |
| | Random forest with PCA | 84.9 | 0.85 | 0.85 | 0.85 |
| | Random Forest with LDA | 67.92 | 0.68 | 0.68 | 0.68 |
| | Random Forest with FSPeSOA | **91.54** | 0.92 | 0.92 | 0.91 |
| | SVM with PCA | 72.64 | 0.73 | 0.73 | 0.71 |
| | SVM with LDA | 62.26 | 0.39 | 0.62 | 0.48 |
| | SVM with FSPeSOA | **91.54** | 0.93 | 0.92 | 0.91 |
| Pima | KNN with PCA | 57.14 | 0.55 | 0.57 | 0.56 |
| | KNN with LDA | 62.33 | 0.67 | 0.62 | 0.63 |
| | KNN with FSPeSOA | 59.74 | 0.58 | 0.60 | 0.59 |
| | Random forest with PCA | 63.63 | 0.61 | 0.64 | 0.62 |
| | Random Forest with LDA | 57.14 | 0.59 | 0.57 | 0.58 |
| | Random Forest with FSPeSOA | 64.28 | 0.64 | 0.64 | 0.64 |
| | SVM with PCA | 67.09 | 0.65 | 0.67 | 0.65 |
| | SVM with LDA | 65.36 | 0.43 | 0.65 | 0.52 |
| | SVM with FSPeSOA | 65.58 | 0.62 | 0.66 | 0.60 |

Table 3b. Comparison based on accuracy, precision, recall and F1-score parameters

| Dataset | Algorithm | Accuracy | Precision | Recall | F1 Score |
|---|---|---|---|---|---|
| Vehicle | KNN with PCA | 43.70 | 0.45 | 0.40 | 0.43 |
| | KNN with LDA | 38.58 | 0.48 | 0.39 | 0.36 |
| | KNN with FSPeSOA | **57.05** | 0.58 | 0.57 | 0.57 |
| | Random forest with PCA | 30.31 | 0.32 | 0.30 | 0.30 |
| | Random Forest with LDA | 35.82 | 0.36 | 0.36 | 0.35 |
| | Random Forest with FSPeSOA | **52.35** | 0.52 | 0.52 | 0.52 |
| | SVM with PCA | 38.18 | 0.30 | 0.38 | 0.33 |
| | SVM with LDA | 30.31 | 0.14 | 0.30 | 0.19 |
| | SVM with FSPeSOA | **58.82** | 0.59 | 0.59 | 0.58 |
| Wine | KNN with PCA | 80 | 0.90 | 0.80 | 0.80 |
| | KNN with LDA | 64.4 | 0.72 | 0.64 | 0.65 |
| | KNN with FSPeSOA | 69.44 | 0.69 | 0.69 | 0.69 |
| | Random forest with PCA | 75.5 | 0.81 | 0.76 | 0.76 |
| | Random Forest with LDA | 71.1 | 0.77 | 0.71 | 0.72 |
| | Random Forest with FSPeSOA | 55.55 | 0.64 | 0.56 | 0.57 |
| | SVM with PCA | 80 | 0.90 | 0.80 | 0.80 |
| | SVM with LDA | 20 | 0.04 | 0.20 | 0.07 |
| | SVM with FSPeSOA | 38.88 | 0.22 | 0.39 | 0.28 |
| Wisconsin | KNN with PCA | 91.81 | 0.92 | 0.92 | 0.92 |
| | KNN with LDA | 63.74 | 0.41 | 0.64 | 0.50 |
| | KNN with FSPeSOA | 79.04 | 0.83 | 0.82 | 0.84 |
| | Random forest with PCA | 90.64 | 0.91 | 0.91 | 0.91 |
| | Random Forest with LDA | 76.6 | 0.76 | 0.77 | 0.76 |
| | Random Forest with FSPeSOA | 84.61 | 0.79 | 0.81 | 0.82 |
| | SVM with PCA | 94.15 | 0.94 | 0.94 | 0.94 |
| | SVM with LDA | 63.74 | 0.41 | 0.64 | 0.50 |
| | SVM with FSPeSOA | **95.61** | 0.96 | 0.96 | 0.96 |

Table 4. Win-loss ratio of algorithms for different datasets

| Dataset | Classifier | Accuracy | Precision | Recall | F1 Score |
|---|---|---|---|---|---|
| Iris | KNN | Win | Win | Win | Win |
|  | Random Forest | Win | Win | Win | Win |
|  | SVM | Win | Win | Win | Win |
| Glass | KNN | Win | Win | Win | Win |
|  | Random Forest | Win | Win | Win | Win |
|  | SVM | Win | Win | Win | Win |
| Ion | KNN | Win | Win | Win | Win |
|  | Random Forest | Win | Win | Win | Win |
|  | SVM | Win | Win | Win | Win |
| Pima | KNN |  |  |  |  |
|  | Random Forest | Win |  | Win | Win |
|  | SVM |  |  |  |  |
| Vehicle | KNN | Win | Win | Win | Win |
|  | Random Forest | Win | Win | Win | Win |
|  | SVM | Win | Win | Win | Win |
| Wine | KNN |  |  |  |  |
|  | Random Forest |  |  |  |  |
|  | SVM |  |  |  |  |
| Wisconsin | KNN |  |  |  |  |
|  | Random Forest |  |  |  |  |
|  | SVM | Win |  | Win | Win |
| Win/Loss |  | 14Win/7Loss | 12Win/9Loss | 14Win/7Loss | 14Win/7Loss |

The above tables 3 and 4 show that our proposed FS-PeSOA algorithm generates most number of "Win"s for supervised feature selection. To summarize in the context of the 7 UCI datasets used in the experiments,

- FS-PeSOA gives better average accuracy than LDA and PCA algorithms.
- FS-PeSOA has a number of "Win"s than any other algorithm (refer table 4).

**5. Conclusion**

In this chapter, a new supervised feature selection approach based on penguins search optimization has been presented. In this chapter, we have summarized the workflow of penguins to an algorithm and tried to use them in a manner that has been beneficial to the task of feature selection for different data sets. In terms of performance of the proposed FS-PeSOA algorithm, experiments using seven publicly available datasets have shown that it has given better results than the full feature set and also the benchmark algorithms of feature selection. It has been compared against PCA and LDA for supervised feature selection. As a future work, this proposed algorithm can be used in science and development purposes and also in biomedical research areas where data analysis is required for the identification of various patterns of different diseases.

**Acknowledgment**

No research funding has been received for this work.